\pdfoutput=1

\documentclass[11pt]{article}

\usepackage{emnlp2021}

\usepackage{times}
\usepackage{latexsym}
\usepackage{epsfig}
\usepackage{graphicx}
\usepackage{amssymb}
\usepackage{amsmath}
\usepackage{multirow}
\usepackage{booktabs}
\usepackage{tabularx}
\usepackage{pifont}
\usepackage[T1]{fontenc}

\usepackage[utf8]{inputenc}

\usepackage{microtype}

\DeclareMathOperator*{\argmax}{\arg\!\max}

\title{Hierarchical BERT for Medical Document Understanding}

\author{Ning Zhang \and  Maciej Jankowski \\
        IQVIA \\ \{ning.zhang, mjankowski@iqvia.com\} }

\begin{document}

\maketitle
\begin{abstract}
Medical document understanding has gained much attention recently. One representative task is the International Classification of Disease (ICD) diagnosis code assignment. Existing work adopts either RNN or CNN as the backbone network because the vanilla BERT cannot handle well long documents (>2000 tokens). One issue shared across all these approaches is that they are over specific to the ICD code assignment task, losing generality to give the whole document-level and sentence-level embedding. As a result, it is not straightforward to direct them to other downstream NLU tasks. Motivated by these observations, we propose Medical Document BERT (MDBERT) for long medical document understanding tasks. MDBERT is not only effective in learning representations at different levels of semantics but efficient in encoding long documents by leveraging a bottom-up hierarchical architecture. Compared to vanilla BERT solutions: 1, MDBERT boosts the performance up to relatively 20\% on the MIMIC-III dataset, making it comparable to current SOTA solutions; 2, it cuts the computational complexity on self-attention modules to less than 1/100. Other than the ICD code assignment, we conduct a variety of other NLU tasks on a large commercial dataset named as TrialTrove, to showcase MDBERT's strength in delivering different levels of semantics.


\end{abstract}

\begin{figure*}
\begin{center}
\includegraphics[width=0.9\linewidth]{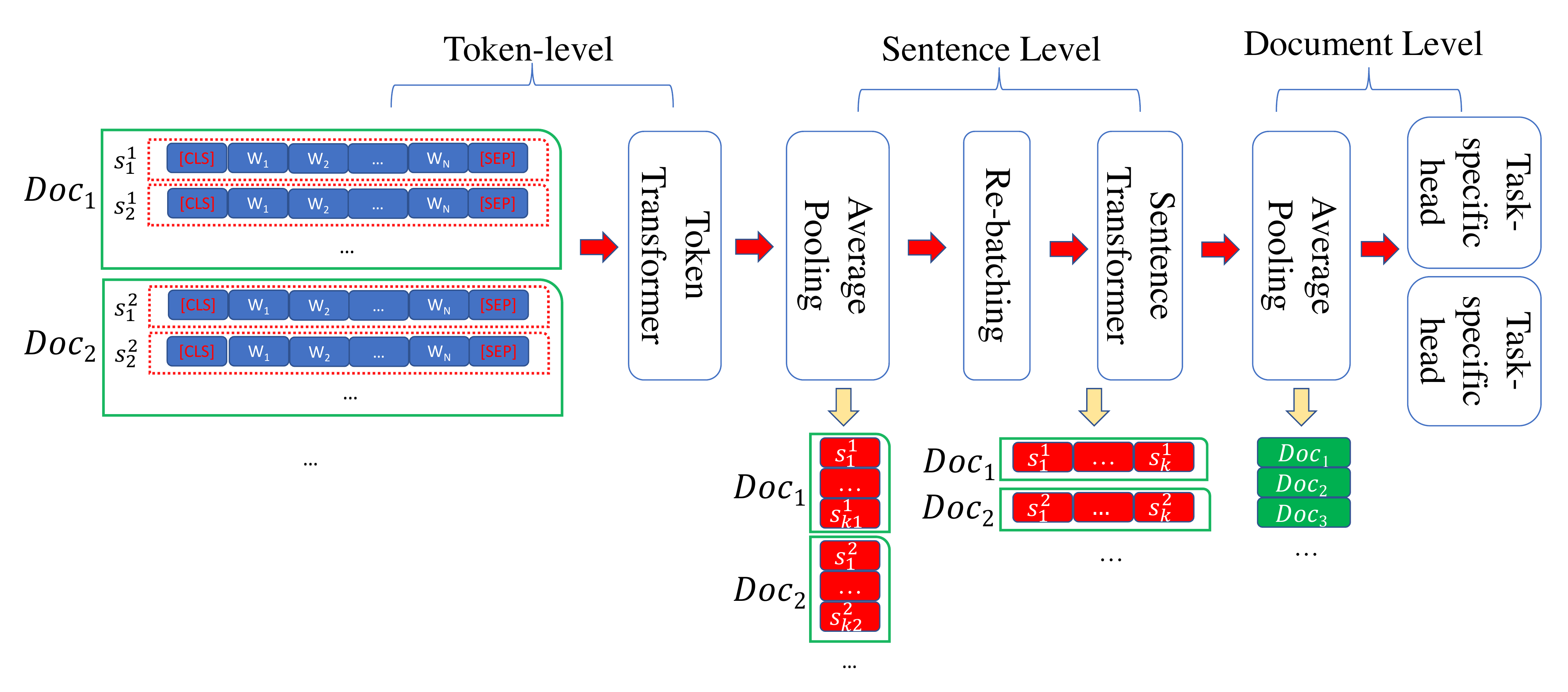}
\end{center}
\caption{The architecture of MDBERT. The input is single sentences from different documents. Sentences that belong to different documents are re-batched in the re-batching module for the upper level of semantics learning. Transformer blocks are working on different levels of semantic. Semantics are aggregated by average pooling in a bottom-up fashion. Note that "Task-Specific Heads" can be attached to any level of semantic. The position of heads is determined by the semantics they are working on.}
\label{fig_MDBERT}
\end{figure*}

\section{Introduction}
Clinical notes are usually generated by physicians and nurses in clinical institutions to record the state of patients. Among all the information the state provides, disease diagnosis, health issues, and related procedures are of great interest \cite{johnson2016mimic, choi2016doctor,avati2018improving}. To represent this information, in a unified way, World Health Organization (WHO) provides a coding system named as International Classification of Diseases (ICD). In this coding system, diseases and procedures are grouped and coded hierarchically. However, assigning proper ICD codes to a clinical note is not easy. It requires strong medical expertise and is error-prone. To reduce the human effort, many approaches leveraging NLP techniques have been devised \cite{mullenbach2018explainable, ji2020medical, li2020icd,vu2020label,cao2020hypercore,xie2019ehr}. 

All these approaches formulate ICD code assignment as a multi-label multi-class problem. They adopt either RNN or CNN as the backbone network followed by a label-wise attention module before doing the classification. To further improve the model, they all leverage the code description to provide extra guidance for training. Unlike the common trend, BERT \cite{ji2020medical, devlin2018bert,vaswani2017attention} based solutions fail to deliver comparable performance. The major cause is over truncation. BERT \cite{vaswani2017attention,lee2020biobert} normally accepts the token sequence that at most be 512 long, while clinical notes easily exceed this limit by a large scale (often be of over 2000 tokens). As a result, over 70\% of the content is lost, and the performance struggles.

Label-wise attention plays a key role in all SOTA approaches \cite{mullenbach2018explainable, ji2020medical, li2020icd,vu2020label,cao2020hypercore,xie2019ehr}. It helps models to save efforts on modeling large context and concentrate on aspects that are critical for classification. However, this attention drives the document embeddings over-specific to single diseases, making them useless for other downstream tasks. Other than that, even being trained on the large textual context, there is no easy way to derive sentence-level embedding from these models. All these facts indicate their limitations when transferred to other understanding tasks.

One concrete scenario is analyzing clinical trial protocols (protocols\footnote{Examples are available in \href{https://www.clinicaltrials.gov/}{ClinicalTrial.gov}}). These documents summarize information about clinical trials of new drugs, devices, and therapies into several aspects, such as target diseases, trial design, patient segments, inclusion and exclusion criteria, sponsors, etc. Assigning ICD code is still an important task when analyzing this type of medical documents. However, additional information such as patient segment classification and recruiting criteria grouping are also important in planning clinical trials. In this case, training dedicated models for each label in each task is overwhelming. A single model that provides general document level embedding is more preferable. 

Moreover, when dealing with clinical protocols, much effort is invested in mapping non-standard disease names to standard ones (ex. MeSH terms or SNOMED terms\footnote{They both are collections of computer processable collection of medical terms}). For instance, given a fuzzy disease name "Brain Aging", a standard disease name "Age-Related Brain Degeneration" is more friendly for clinical researchers to later group the trials. To this end, a quick solution is to resort to the phrase- or sentence-level embedding. In previous work \cite{mullenbach2018explainable, cao2020hypercore}, this type of embedding is given by a separate network and tightly coupled with labels. The main network provides no mechanism to calculate general-purpose sentence-level embedding.


Motivated by these facts, we propose a hierarchical BERT named as Medical Document BERT (MDBERT, Fig. \ref{fig_MDBERT}). This model works on different levels of semantics simultaneously from the token to sentence and eventually to document. Training and testing this model is end-to-end. Task- and level-specific supervision can be easily injected at different places. With this design, MDBERT embraces multi-instance learning in nature and can be easily adapted to different understanding tasks. 

Our key contributions are summarized as below: \\
1, we propose Medical Document BERT (MDBERT), a hierarchical BERT for ICD code assignment. To the best of our knowledge, this is the first BERT based approach that performs comparably to SOTA solutions for the ICD code assignment task on the MIMIC-III dataset \cite{johnson2016mimic}.\\
2, we provide a unified way for code description augmentation without maintaining a separate network for ICD code embedding. \\
3, we provide an easy way to derive sentence- and document-level embedding from MDBERT that can fit directly to other downstream understanding tasks. By doing so, we demonstrate its generality on long document understanding tasks.

\section{Related Work}

\subsection{ICD Code Assignment}
ICD code assignment has received much attention recently. Mainstream approaches consist of three main components: 1, backbone network to calculate contextualized embedding; 2, head network equipped with label-wise attention to derive constant length embedding for each class; 3, code description augmentation for unseen classes in the training set as well as implying regularization on representation learning.

More concretely, CAML \cite{mullenbach2018explainable}, MultiResCNN \cite{li2020icd} and  GatedCNN-NCI \cite{ji2020medical} adopt CNN as the backbone network. MultiResCNN adopts the multi-filter strategy to strengthen the backbone network while GatedCNN-NCI applies gates to control the information flows into the head network. Different from the aforementioned work, Vu et. al \citeyear{vu2020label} use LSTM as the backbone and achieve the current best performance on MIMIC-III. All these approaches exploit label-wise attention to calculate representations specific to each class. CAML and MultiCNN apply the vanilla form of attention which can be summarized as a single linear layer while Vu et al. take a more structured way following \cite{lin2017structured}. This structured attention also can be abstracted as a two-layer network. 

ICD code descriptions provide rich guidance for the whole model training. Often a separate network is in use to encode these descriptions \cite{cao2020hypercore, mullenbach2018explainable}. The code embeddings will then interact with the main network during training. The interaction mechanism differs across different approaches. In CAML, code embeddings simply regularize the classifier weights in a Euclidean distance form. HyperCore \cite{cao2020hypercore} trains another graph convolutional network to explicitly exploit the code hierarchy. The outcome code embeddings act as attention weights for document representation learning. Similar to HyperCore, the code hierarchy information is also exploited in \cite{vu2020label,ji2020medical,xie2019ehr}.

\subsection{BERT on Sentence and Document Semantics}
Sentence and Document embedding aim to encode sentences and documents as constant length vectors. This research area has been very active for a long time. Thoroughly reviewing related work is out of the scope of this paper. We only briefly discuss ones that are closely related to our paper. 

In vanilla BERT, sentence level embedding is represented by the special token "[CLS]" at the last Transformer layer \cite{zhang2019bertscore, may2019measuring}. This has been a standard way for a while. Sentence BERT (SBERT \citeauthor{reimers2019sentence}), however, demonstrates that this approach is sub-optimal. Instead, SBERT calculates sentence-level embedding by average pooling. We adopt this idea in our approach when encoding sentences. 

To encode long documents, HiBERT \cite{zhang2019hibert} proposed a hierarchical BERT. It is of an encode-decoder architecture and equipped with sentence-level Masked Language Model objective. In the encoder, it still uses the contextualized embedding from a single special token to represent a sentence. In the decoder, masked sentences will be predicted following the MLM framework. In our work, we keep it simple by removing this unsupervised sequence-to-sequence MLM. Instead, we simply apply average pooling to calculate the document embedding. Therefore, our model is trained fully supervisedly.

\section{Algorithm}
\subsection{Architecture}
The whole architecture of our Medical Document BERT is illustrated in Fig. \ref{fig_MDBERT}. Our MDBERT is fully Transformer \cite{vaswani2017attention} based. We use the Transformer Encoder layer as the building block. Transformer blocks are working on different levels of semantics. Note that, in our definition, the semantic level of an embedding is determined by the entity it is attached to. A transformer provides context while not change the semantic level. Starting from the token level as vanilla BERT \cite{devlin2018bert}, our model aggregates semantics to the sentence and document level by average pooling in a bottom-up fashion. Sentence and document embeddings can be easily derived at the average pooling. The input of MDBERT is individual sentences coming from different documents. Re-Batching aggregates all sentences that belong to the same document and reformulate each document as a sequence of sentences. This strategy helps much in saving memory usage and accelerating the training process.

Task-specific heads can be attached to any level of semantics. The position is fully determined by the task itself. For instance, the vanilla Masked LM objective works at the token level (contextualized). Therefore, this head can be attached to the last layer of the token transformer. Sentence level NLU tasks such as ones defined in SentEval \cite{conneau2018senteval} can leverage embeddings generated after the first averaging pooling \cite{reimers2019sentence}. In our work, we wrap up classifier weights with the attention mechanism as the task-specific head. Therefore, no special attention mechanism is depicted in the backbone network. We will discuss the attention mechanism in MDBERT later.

For the ICD Code Assignment task, we can attach the multi-class multi-label classifier either before or after the second average pooling. On the MIMIC-III dataset, we adopt the former option. In this way, by our definition, our ICD code assignment head is still working with sentence level semantics. Aggregation of semantics is conducted by the head itself instead of the backbone network. In contrast, in our other series of experiments on the TrialTrove dataset, our classifier heads are all working with document level of semantics produced by the second average pooling.

\begin{figure}[t]
    \centering
    \includegraphics[width=1.0\linewidth]{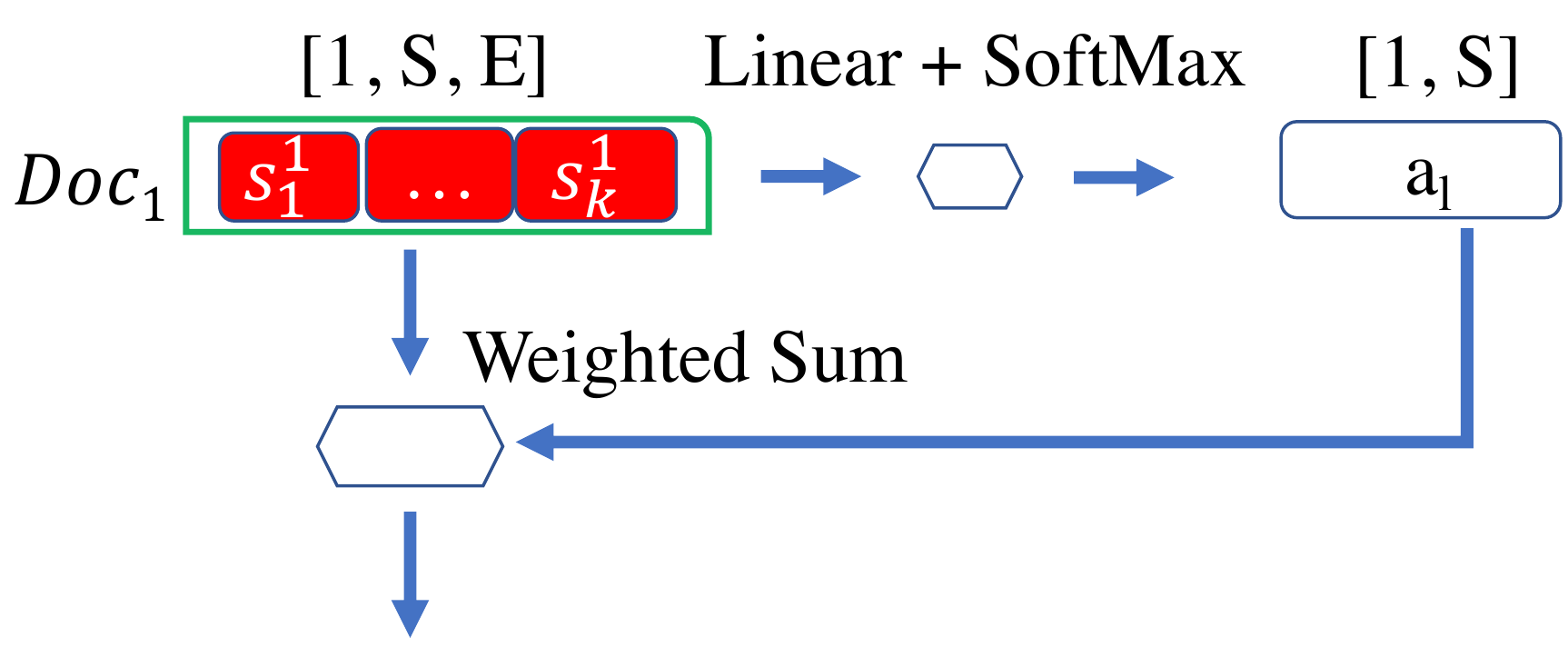}
    \caption{Label-wise attention}
    \label{fig_attention}
\end{figure}

\subsection{Complexity}
By breaking the document into separate sentences, MDBERT cuts down the computational complexity at a great scale compared to vanilla BERT implementations. Take a document of 512 tokens as an example and assume it consists of 16 sentences. In common practice, no matter whether or not sentences are separated by "[SEP]", these 512 tokens are fed as a single sequence to BERT, making the complexity of the self-attention be $\mathcal{O}(n^2\cdot d)$, where $n$ and $d$ denote the token number and hidden dimension. However, in MDBERT, the complexity is cut to $\mathcal{O}((n/s)^2\cdot d \cdot s) = \mathcal{O}(n^2\cdot d / s)$, where $s$ is the number of sentences. If we take in the sentence transformer blocks and assume they are of the same depth as the token transformer. The complexity cost by self-attention is refined as $\mathcal{O}(n^2\cdot d/s + s^2\cdot d)$. In this particular example where $s=16$, the computational complexity on self-attention is cut to $\sim$1/16. 

For long documents, the analysis of complexity above holds for Multi-Heads Self-Attention (MHSA) modules. As we know, MHSA is of the complexity $\mathcal{O}(n^2\cdot d + n\cdot d^2)$. Normally $\mathcal{O}(n\cdot d^2)$ dominates because $n$ is usually much smaller than $d$. However, when dealing with long documents (Tab. \ref{tab_doc_statistics}) where $n$ can easily grows much larger than $d$, the $\mathcal{O}(n^2\cdot d)$  term becomes to the major cost. Then we come back to our analysis on Self-Attention. Moreover, $s$ grows linearly to $n$ in nature, making the cut-off larger as the document length grows.

\subsection{Label-Wise Attention}
Label-wise attention relieves the model from learning the very complex context.
In MIMIC-III, documents are overwhelmingly long. After the standard pre-processing proposed by Mullenbach et al. \citeyear{mullenbach2018explainable}, the average token number of each document is roughly 1500. Long documents provide much redundant information while the task is specific. To fill the gap, label-wise attention drives the model to focus on aspects specific to each class. In our model, we apply the vanilla form of attention used in CAML \cite{mullenbach2018explainable}. This attention is working with sentence level embedding followed by the logistic regression. 

More specifically (Fig. \ref{fig_attention}), given the representation $\mathbf{H}^i \in \mathbb{R}^{S\times E}$ for document $i$ which has $S$ sequences and each sentence is encoded as an $E$ dimensional vector, the attention will give attentive weights $\mathbf{a}_l \in \mathbb{R}^S$ for class $l$ on each sentence:
\begin{equation}
\label{eq_attn}
\mathbf{a}_l = SoftMax(\mathbf{H}^i \mathbf{w}_l),
\end{equation}
where $\mathbf{w}_l \in \mathbb{R}^E$ denotes the attention weights for class $l$ and $SoftMax(\mathbf{x})= \frac{exp(\mathbf{x})}{\sum{exp(x_i)}}$. The attentive weights are then used to summarize all sentence embeddings of document $i$ to a constant length vector $\mathbf{u}_l^i$ for class $l$: 
\begin{equation}
\label{eq_wsum}
\mathbf{u}_l^i =\mathbf{a}_l^{\top}\mathbf{H}^i .
\end{equation}
The final score on class $l$ is produced by another linear layer:
\begin{equation}
\label{eq_class}
\hat{y}_l^i = \sigma(\mathbf{v}_l^{\top} \mathbf{u}_l^i + b_l),
\end{equation}
where $\{\mathbf{v}_l; b_l\}$ are the classifier parameters.

\subsection{Augmentation with ICD CODE Descriptions}
In MIMIC-III, there are codes only appearing once, making neither the training set nor testing set see the full set of the codes. This lead people to resort to code descriptions for augmentation. In most of the previous work \cite{mullenbach2018explainable, cao2020hypercore, ji2020medical}, separate networks are trained to encode these descriptions. Code embeddings will interact with document representation (code-specific) or classifier parameters to provide additional supervision. Interactions are of different types, ranging from simply as a euclidean distance to complicated graph message passing. 

In our solution, we simply take each code description as a single sentence document and train the network to correctly predict the one-hot label. By doing so, the code descriptions simply serve as data augmentation and we can save the efforts on maintaining a separate network.

\subsection{Training Loss}
Given the sparseness of positive samples with respect to each class, we slightly change the form of commonly used cross entropy loss by adding more weights on positive samples: 
\begin{equation}
\label{eq_weighted_CE}
\mathcal{L} = -\sum_{l}[w_{l}y_{l}log(p_l) + (1-y_l)log(1-p_l)] ,
\end{equation}
where $w_l \geq 1$ and $l$ denote the positive sample weight and $l$th class. In our experiments, $w_l$ is kept the same across different classes. 
However, we can also differentiate $w_l$ to control the balance across different classes. In our experiments, $w_l$ is typically set in [1, 5]. Both the documents and code descriptions adopt this loss form and the final loss is: 
\begin{equation}
\label{eq_loss}
\mathcal{L} = \mathcal{L}_{Doc} + \mathcal{L}_{Desc}.
\end{equation}

\subsection{Sentence and Document Embedding}
If label-wise attention is adopted, we still have no single generic embedding to use for other document level NLU tasks. In the meantime, deriving sentence level embeddings from existing approaches is not straightforward and untested. To this end, we demonstrate that through average pooling operations (as shown in Fig.\ref{fig_MDBERT}), we can easily acquire sentence and document level semantics for other understanding tasks. All these tasks are conducted on the commercial clinical trial dataset TrialTrove\footnote{Also referred to as Citeline\textsuperscript{\textregistered} TrialTrove in some other places, provided by \href{https://pharmaintelligence.informa.com/products-and-services/data-and-analysis/trialtrove}{Informa}.}.
\begin{table}[t]
  \centering
  \resizebox{\columnwidth}{!}{%
    \begin{tabular}{l|cc|cc|cc}
    \hline
    \multirow{2}{2em}{Dataset} 	& \multicolumn{2}{c}{Sents/Doc} & \multicolumn{2}{c}{Tokens/Sent} &  \multicolumn{2}{c}{Tokens/Doc} \\ 
    & Mean & STD  & Mean & STD & Mean & STD \\ 
    MIMIC-III & 106.4  & 54.2  &  19.3 & 22.8 & 2,187.7& 1,166.5 \\
    TrialTrove & 9.6  & 8.6 & 34.5 &  35.2 &339.6 & 328.5 \\ \hline
    \end{tabular}
    }
    \caption{Document statistics on sentences and tokens after pre-processing. Note that the statistics on tokens (not integral words) are determined by the BioBERT tokenizer.}
  \label{tab_doc_statistics}
\end{table}

\begin{table}[t]
  \centering
  \resizebox{\columnwidth}{!}{%
    \begin{tabular}{lcccc}
    \hline
    Task & train & test &  \#classes & \#labels \\ \hline
    ICD10 Code & 222,148  & 55,473  &  725 & 5.60 \\ \hline
    Patient Segment & 202,790  & 50,599 & 660 & 3.25 \\ \hline
    Age Group & 196,435 & 48,827  & 5 & 1.91 \\ \hline
    Gender & 207,409 & 51,513   & 3 & 1\\ \hline
    Combined & 266,236 & 66,559 & 1,393 & 9.33\\ \hline
    \end{tabular}
    }
    \caption{Label Statistics of TrialTrove understanding tasks. Samples are marked as valid as long as one positive label is attached to that sample. Otherwise, it will be ignored for the specific task.}
  \label{tab_trialtrove_statistics}
\end{table}

\begin{table*}[h!]
\small
\centering
\setlength{\tabcolsep}{2pt}
\resizebox{2\columnwidth}{!}{
\begin{tabular}{lcc|cc|c|cc|cc|cc}
\hline
 \multirow{3}{4em}{Model} 	& \multicolumn{5}{c}{MIMIC-III Top-50 Codes} & \multicolumn{5}{c}{MIMIC-III Full Codes} \\
	\cline{2-12} 
	& \multicolumn{2}{c}{AUC-ROC} & \multicolumn{2}{c}{ F1 } & {P@K} & \multicolumn{2}{c}{AUC-ROC} & \multicolumn{2}{c}{ F1 } & \multicolumn{2}{c}{P@K}\\  
 	&Macro &Micro& Macro&Micro & P@5 &Macro &Micro& Macro&Micro & P@8& P@15\\
\midrule	
DR-CAML~\cite{mullenbach2018explainable}	&	88.4	&	91.6	&	57.6	&	63.3	&	61.8	&	89.7	&	98.5	&	8.6	&	52.9	&	69.0 & 54.8	\\
LEAM~\cite{wang2018joint}	&	88.1	&	91.2	&	54.0	&	61.9	&	61.2	&	-	&	-	&	-	&	-	&	- & - 	\\
MSATT-KG \cite{xie2019ehr}  & 91.4  &93.6  & \textit{63.8} & 68.4  & 64.4 & 91.0 & \textbf{99.2} &9.0 &55.3 &72.8 &58.1	\\ 
MultiResCNN~ \cite{li2020icd}		&	89.9	&	92.8	&	60.6	&	67.0	&	64.1	& 91.0 &	98.6	&	8.5	&	55.2	&	73.4 & \textbf{58.4}	\\
HyperCore~\cite{cao2020hypercore}	&	89.5 	&	92.9	&	60.9	&	66.3	&	63.2 	&	\textbf{93.0}	&	\textit{98.9}	&	9.0	&	55.1	&	73.2 & 	57.9\\
GatedCNN-NCI \cite{ji2020medical}	& \textit{91.5}	&	\textit{93.8}	&	62.9	&	\textit{68.6}	&	\textit{65.3}		&	\textit{92.2}	&	\textit{98.9} &	\textit{9.2}	&	\textit{56.3}	&	\textbf{73.6} & -\\
JointLAAT \cite{vu2020label}	& \textbf{92.5}	&	\textbf{94.6}	&	\textbf{66.1}	&	\textbf{71.6}	&	\textbf{67.1}		&	92.1	&	98.8 &	\textbf{10.7}	&	\textbf{57.5}	&	\textit{73.5} & \textbf{59.0} \\ \hline
BERT-Base \cite{ji2020medical}	& 80.6	&	85.2	&	43.3	&	53.2	&	53.3		&	-	&	- &	-	&	-	&	- & - \\ 
CLinicalBERT \cite{ji2020medical}	& 81.0	&	85.6	&	43.9	&	54.3	&	54.5 &	-	&	- &	-	&	-	&	- & - \\ 
BERT-ICD \cite{pascual2021towards}	& 84.5	&	88.6	&	-	& -	&	- &	-	&	- &	-	&	-	&	- & - \\ 
MDBERT-SBERT (Ours) 	& 91.1	&	93.1	&	64.4	&	68.1	&	64.3 &	-	&	- &	-	&	-	& - & - \\ 
MDBERT (Ours)	& {91.8}	&	93.6	&	{65.9}	&	{69.2}	&	{65.4} & 92.5 &	98.9	&10.1	&55.5	&72.7	&57.7 \\ \hline


MDBERT+avg (Ours)	& \textbf{92.8}	&	\textbf{94.6}	&	\textbf{67.2}	&	\textbf{71.7}	&	\textbf{67.4} &	\textbf{94.2}	&	\textbf{99.2}   &	\textit{10.4}	&	\textbf{57.6}	& \textbf{75.0} & \textbf{59.6} \\ \hline


\end{tabular}
}
\caption{Results on MIMIC-III with top-50 and full codes. \textbf{Bold} text and \textit{italic} text denote the best and second best for both the previous and current records. "LATT" stands for label attention. "MDBERT-SBERT" means the sentence transformer is removed. "+avg" means we simply apply the model averaging technique on three trained models.}
\label{tab_mimic_iii}
\end{table*}

\section{Experiment}

\subsection{Datasets}
We evaluate our MDBERT in two datasets: MIMIC-III\cite{johnson2016mimic} and \href{https://pharmaintelligence.informa.com/products-and-services/data-and-analysis/trialtrove}{TrialTrove}.

\textbf{MIMIC-III}, Medical Information Mart for Intensive Care III, is a public dataset, containing text and structured records of patients admitted to the Intensive Care Unit (ICU). We focus on predicting ICD9 codes (version 9 of ICD code) from discharge summaries. The full dataset contains 52,722 discharge summaries and 8,929 unique codes in total. Following \cite{mullenbach2018explainable}, the full dataset is split into train (47,719), dev (1,631) and test (3,372) sets. Other than the full set, a subset is also extracted only focusing on the 50 most frequent codes. This results in 8,066, 1,573, 1,729 samples for train, validation, and test sets respectively.

\textbf{TrialTrove} is a commercial dataset produced by Informa. It collects pharmaceutical clinical trial data from different sources. It summarizes important information such as trial title, trial objective, trial design, target diseases (indications), patient segments, inclusion and exclusion criteria, etc. On this dataset, we mainly focus on textual information and formulate understanding tasks as follows:
1, using the trial title, objective, patient population, and inclusion criteria predict frequent ICD10 (10th version of ICD) codes, patient segment codes, patient age groups, and gender; 2, indication standardization: given a non-standard disease name finding the closest standard disease name (SNOMED). In task 1, all sub-tasks are multi-label, multi-class classification except for the gender prediction. TrialTrove is keeping updating. At the date when we conducted our experiments, we extracted 332,795 trials that at least contains one valid label. 

\subsection{Pre-processing}
For both datasets, documents are first segmented into sentences leveraging the scispacy library. MIMIC-III documents normally contain more than 100 sentences (Tab. \ref{tab_doc_statistics}). After sentence segmentation, we follow the pipeline proposed by \cite{mullenbach2018explainable}: numeric and de-identified tokens removal, ICD code reformation, etc. 

However, for the TrialTrove dataset, we adopt a slightly different process after the segmentation. For document understanding tasks, we concatenate textual information from 4 different fields including trial title, trial objective, patient population, and inclusion criteria. Combinations of these 4 fields differ across different experiment settings. When splitting the data, we only randomly split the dataset into the train(80\%) and test(20\%) set. 

All labels are ensured to exist in both sets. Statistics of label with respect to different tasks are summarized in Tab. \ref{tab_trialtrove_statistics}. Note that, in multi-label multi-class tasks, we only consider labels that occur at least 50 times. For the ICD10 code, we further only consider code at level 4 (one digit after the dot). For age group, we divide human into 5 groups: infants (0-1.5 y/o], children (1.5-12 y/o], teenagers (12-18 y/o], adults (18 -65y/o], seniors (>=65 y/o).

\subsection{Training}
Our model is implemented using Pytorch and HuggingFace libraries. The token transformer is initialized with pre-trained BioBERT\footnote{available at \href{https://github.com/dmis-lab/biobert}{https://github.com/dmis-lab/biobert}} \cite{lee2020biobert}. The whole training consists of two steps: 1, BioBERT is frozen and only layers on top of it are tuned; 2, whole model tuning. The first step normally takes 3 epochs. We adopt AdamW as the optimizer and important hyper-parameters are summarized in Tab. \ref{tab_hyperparmeter}. We apply a similar early stopping strategy as \cite{mullenbach2018explainable} for both datasets. The stop criteria is performance on dev set (test set for TrialTrove) stops improving for 3 consecutive epochs. This usually requires 7-12 epochs of whole model tuning on both datasets. 

\subsection{Evaluation Metrics}
On MIMIC-III, we follow the standard evaluation protocol proposed by \cite{mullenbach2018explainable}, including AUC score, F1-score, and Precision@X (X differs for different task settings). Both Micro and Macro averaging strategies are applied in this multi-label multi-class problem. The former does not differentiate different classes and all scores are calculated as standard binary classification problems while the latter calculates class-specific metrics before averaging the scores. Therefore, Micro puts more weight on frequent classes while Macro takes all classes equally.

On TrialTrove, we use the AUC score to evaluate the indication standardization task. The input is raw disease names in trial documents and ground truth are the SNOMED names attached to it. Note that each disease name can have multiple SNOMED names attached. We only consider the Top1 hit. The Top1 hit is marked as true as long as it is in the ground truth SNOMED set. In addition to the metrics we used for MIMIC-III, we also adopt Macro Average Precision and Accuracy to evaluate multi-label and single-label multi-class problems respectively for document level understand tasks. Macro AP score is defined as: 
\begin{equation}
\label{eq_ap}
AP_{Macro} = (1/L) \sum^{L}_{l} \sum_{i} (R_i^l-R_{i-1}^l)P_i^l,
\end{equation}
where $P_i^l$, $R_i^l$, $L$ denote the precision and recall at the i-th threshold and number of classes respectively.

\begin{table}[t]
  \centering
  \resizebox{\columnwidth}{!}{%
    \begin{tabular}{lccccc}
    \hline
    Dataset & \#layer & \#Sent & \#Token & LR & $w_l$\\ \hline
    MIMIC-III-top50 & 4  & 256  & 62 & 1e-5 & \{1.5, 3\}\\ \hline
    MIMIC-III-full & 6  & 256  & 62 & 1e-5 & \{3, 5\}\\ \hline
    TrialTrove & 4  & 32  & 62 & 1e-5 & \{1, 1\} \\ \hline
    \end{tabular}
    }
    \caption{Hyper-parameters used during training. \#layer indicates the depth of Sentence Transformer. Note that \#Sent and \#Token denote the number of sentences that one document and the maximum length that allow for one single sentence. We use different $w_l$ for $\mathcal{L}_{Doc}$ and $\mathcal{L}_{Desc}$ in Eq. \ref{eq_loss}. Therefore, $w_l$ is listed separately as a pair.}
  \label{tab_hyperparmeter}
\end{table}

\begin{table}[t]
  \centering
  \resizebox{\columnwidth}{!}{%
    \begin{tabular}{ccc|cc|cc|c}
    \hline
    \multirow{2}{2em}{Sent \\ Trans} &\multirow{2}{2em}{Label \\ Attn} &  \multirow{2}{2em}{Desc \\ Aug} & \multicolumn{2}{c}{AUC-ROC} & \multicolumn{2}{c}{F1} & \multirow{2}{2em}{P@5} \\ 
    & & & Macro & Micro & Macro & Micro & \\ 
    - & \checkmark & \checkmark   & 91.1  & 93.1 & 64.4 & 68.1 & 64.3 \\ 
    \checkmark& \checkmark & -  & 91.1  & 93.0 & 63.2 & 67.2 & 64.4 \\ 
    \checkmark &- & \checkmark  & 90.0  & 92.1 & 62.4 & 65.8 & 63.6\\ 
    \checkmark & \checkmark & \checkmark  & 91.8  & 93.6 & 65.9 & 69.2 & 65.3  \\ \hline
    \end{tabular}
    }
    \caption{Ablation study of MDBERT on MIMIC-III Top-50 codes. "Sent Trans", "Label Attn", "Desc Aug" stand for sentence transformer, label attention, and code description augmentation respectively. When the sentence transformer is removed, MDBERT falls to be MDBERT-SBERT in Tab. \ref{tab_mimic_iii}. }
  \label{tab_ablation_mimic3}
\end{table}


\subsection{ICD9 Code Assignment on MIMIC-III}
We tested our MDBERT on the MIMIC-III dataset for both the full label and top-50 label setting. Results are summarized in Tab. \ref{tab_mimic_iii}. As we can easily tell that compared to SOTA solutions, the single model performance of MDBERT always ranked at the second or third best, making it a very competitive solution. However, to our surprise, we find models trained separately perform well in different classes. Motivated by this, we average the predictions of 3 models that trained independently. The performance improves at a large margin. We conjecture the main reason lies in the relatively small dataset size, making sentence transformer under-trained. 

Compared with previous BERT based models, our model improves the performance at a very large margin. We also trained an MDBERT-SBERT by removing the sentence transformer in Fig. \ref{fig_MDBERT}. The major difference between our MDBERT-SBERT and the SBERT from \cite{ji2020medical} is that they feed a document as a single sequence while we break it into multiple sequences (sentences).

To study the effects brought by the major components of MDBERT, we conduct an ablation study on the top-50 setting. From Tab. \ref{tab_ablation_mimic3} we can see all these modules contribute to the performance while label attention is most critical. Once it is removed, we use the average pooling instead. The performance decreases drastically. Another interesting finding is that performance did not drop much when the sentence transformer is removed. These two findings give us a clue that ICD code assignment may work without fully understanding the rich context given by the long document. 

\begin{table}[t]
  \centering
  \resizebox{\columnwidth}{!}{%
    \begin{tabular}{lccc}
    \hline
    Model & Input Data & \{\#Sen, \#Seq\}& AP \\ 
    SBERT & title + objective & - & 82.0 \\ 
    MDBERT & title + objective & \{16, 62\} & \textbf{83.2} \\ \hline
    MDBERT & all & \{32, 62\} & 85.1 \\ 
    MDBERT+MT & all & \{32, 62\} & \textbf{87.9} \\ \hline
    \end{tabular}
    }
    \caption{ICD10 code prediction on TrialTrove. \{\#Sen, \#Seq\} denotes the max number of sentences that a document is allowed and the max sequence length of a single sentence. "ALL" means all 4 textual fields of interest are used. "MT" indicates multi-task training.}
  \label{tab_icd_trialtrove}
\end{table}

\subsection{Understanding Tasks on TrialTrove}
We conduct experiments on the TrialTrove dataset to mainly answer the following questions: 1, can MDBERT learn document level semantics that can easily be adapted to different understanding tasks? 2, can the model also produce good sentence embedding without introducing extra layers or sentence level supervision? To this end, unless exceptional announcements are made, all label-wise attention is removed from this series of experiments. Representations are all coming from average pooling: sentence level for the indication standardization task; document level for all the document level understanding tasks.

\textbf{ICD10 Code Assignment} As a continual work of ICD code assignment, we also test MDBERT on TrialTrove. We trained an SBERT \cite{reimers2019sentence, ji2020medical} on this dataset as our baseline model. Documents are retained as single sequences with each sentence be separated by the token "[SEP]". To remove the impact of document truncation, we first only combine the trial title and trial objective as the input. All the resulting documents are short (<0.5\% get truncated). In this way, we can better compare the effectiveness of sentence transformer blocks. To make the results concise, we adopt standard Macro Average Precision as the evaluation metric. As we can easily observe from Tab. \ref{tab_icd_trialtrove} that with the similar input information. MDBERT outperforms SBERT at a considerable margin. And with increasingly more data pulled in, the performance improves accordingly.

\begin{table}[t]
  \centering
  \resizebox{\columnwidth}{!}{%
    \begin{tabular}{l|cc|cc|c|c}
    \hline
    \multirow{2}{2em}{Task} & \multicolumn{2}{c}{AUC-ROC} & \multicolumn{2}{c}{F1} & \multirow{2}{2em}{P@8} & \multirow{2}{2em}{AP / \\Acc} \\ 
    & Macro & Micro & Macro & Micro &  & \\ 
    Gender  & -  & - & - & - & -  & 83.2 \\ 
    Age Group  & 98.4  & 99.5 & 92.7 & 97.3 & 38.0  & 94.9  \\
    Patient Seg & 99.8  & 99.8 & 73.4 & 77.1 & 31.1  & 79.2\\ 
    ICD10 & 99.6  & 99.8 & 83.5 & 87.9 & 41.6  & 87.9\\  \hline
    \end{tabular}
    }
    \caption{All document understanding tasks on the TrialTrove. As Gender Classification is a single-label task, we only report accuracy for it. }
  \label{tab_doctask_trialtrove}
\end{table}

\textbf{Will Document Embedding generalize well to different tasks?} As we have already witnessed from Tab. \ref{tab_icd_trialtrove} that multi-task learning does help to improve the performance on ICD10 code prediction. We now summarize the performance of MDBERT on other tasks. As we can easily tell from Tab. \ref{tab_doctask_trialtrove}, the unified document embedding performs very well on all classification tasks.

\textbf{Will MDBERT perform well in sentence embedding?} All the training so far involves no direct sentence supervision. However, by design, MDBERT implicitly learns to encode sentences. To test its effectiveness on encoding sentences, we adopt the indication standardization task: give a raw disease name in the dataset as the query, we are to find the best matching from a subset of the SNOMED database. This subset is generated by collecting 729 most frequent SNOMED diseases in clinical trials and their synonyms from the SNOMED database. The resulting search base contains 3,955 indications. We extract 207 unique raw disease names from TrialTrove as the query terms. By comparing against the ground truth provided by TrialTrove, we evaluate the performance.

We encode all disease names using the MDBERT model trained with multi-task learning. Embeddings are from the first average pooling. We calculate cosine similarity for each query and candidate pair to find the best matching SNOMED indication. We evaluate the AUC score of the top1 hit. By comparing with various baseline approaches (Tab. \ref{tab_indication_stardardization}), we find our MDBERT performs the best. Note that, to give a powerful TF-IDF baseline, we leverage a mature industry solution for search engine ElasticSearch\footnote{https://www.elastic.co/} (ES). We create an ES index using the SNOMED indications and search the indication using the raw disease name. More details can be found in the appendix.

\begin{table}[t]
  \centering
  \resizebox{\columnwidth}{!}{%
    \begin{tabular}{lcc}
    \hline
    Model & contextualized & AUC \\ 
    TF-IDF (BM25)$\dag$ & - & 89.1 \\ 
    BioBERT \cite{lee2020biobert} & \ding{55} & 94.5 \\ 
    BioBERT-large \cite{lee2020biobert} & \ding{55} & 92.5 \\ 
    BioWordVec \cite{zhang2019biowordvec} & \ding{55} & 95.6 \\ \hline
    BioBERT-CLS token \cite{lee2020biobert} & \checkmark & 92.6 \\
    BioBERT-large-CLS token \cite{lee2020biobert} & \checkmark & 88.4 \\
    SBERT (ours) & \checkmark & 95.3 \\ 
    MDBERT (ours) & \checkmark & \textbf{96.8} \\ \hline
    \end{tabular}
    }
    \caption{Indication Standardization on TrialTrove dataset. The SBERT here is from Tab. \ref{tab_icd_trialtrove}. "Contextualized" indicates whether embedding would propagate through a model. TF-IDF is implemented with the ElasticSearch software fully relying on string similarity. "*-CLS token" use the "[CLS]" token embedding from the last hidden layer of BERT.}
  \label{tab_indication_stardardization}
\end{table}

\section{Conclusion}
MDBERT is a powerful solution for medical document understanding tasks. It, by design, fits well with multi-instance learning and affords full potentials to evolve with the training data. It can encode long documents without suffering from the sequence length limitation posed by vanilla BERT. More than that, different levels of semantics are learned end-to-end within a unified fashion. Compared to naive BERT implementation, MDBERT is computationally efficient. 

Other than the good aspects, we find MDBERT is overfitting-prone. Performance stops improving normally after 10 epochs especially when trained on small datasets. This is far fewer than the epochs reported from other work \cite{mullenbach2018explainable, ji2020medical, li2020icd} where the training process can easily last for over 200 epochs. Sentence level of constraints should help such as the MLM mechanism proposed in \cite{zhang2019hibert}. We leave it as our future work.


\bibliography{anthology,custom}

\begin{thebibliography}{21}
\expandafter\ifx\csname natexlab\endcsname\relax\def\natexlab#1{#1}\fi

\bibitem[{Avati et~al.(2018)Avati, Jung, Harman, Downing, Ng, and
  Shah}]{avati2018improving}
Anand Avati, Kenneth Jung, Stephanie Harman, Lance Downing, Andrew Ng, and
  Nigam~H Shah. 2018.
\newblock Improving palliative care with deep learning.
\newblock \emph{BMC medical informatics and decision making}, 18(4):55--64.

\bibitem[{Cao et~al.(2020)Cao, Chen, Liu, Zhao, Liu, and
  Chong}]{cao2020hypercore}
Pengfei Cao, Yubo Chen, Kang Liu, Jun Zhao, Shengping Liu, and Weifeng Chong.
  2020.
\newblock Hypercore: Hyperbolic and co-graph representation for automatic icd
  coding.
\newblock In \emph{Proceedings of the 58th Annual Meeting of the Association
  for Computational Linguistics}, pages 3105--3114.

\bibitem[{Choi et~al.(2016)Choi, Bahadori, Schuetz, Stewart, and
  Sun}]{choi2016doctor}
Edward Choi, Mohammad~Taha Bahadori, Andy Schuetz, Walter~F Stewart, and Jimeng
  Sun. 2016.
\newblock Doctor ai: Predicting clinical events via recurrent neural networks.
\newblock In \emph{Machine learning for healthcare conference}, pages 301--318.
  PMLR.

\bibitem[{Conneau and Kiela(2018)}]{conneau2018senteval}
Alexis Conneau and Douwe Kiela. 2018.
\newblock Senteval: An evaluation toolkit for universal sentence
  representations.
\newblock \emph{arXiv preprint arXiv:1803.05449}.

\bibitem[{Devlin et~al.(2018)Devlin, Chang, Lee, and
  Toutanova}]{devlin2018bert}
Jacob Devlin, Ming-Wei Chang, Kenton Lee, and Kristina Toutanova. 2018.
\newblock Bert: Pre-training of deep bidirectional transformers for language
  understanding.
\newblock \emph{arXiv preprint arXiv:1810.04805}.

\bibitem[{Ji et~al.(2020)Ji, Pan, and Marttinen}]{ji2020medical}
Shaoxiong Ji, Shirui Pan, and Pekka Marttinen. 2020.
\newblock Medical code assignment with gated convolution and note-code
  interaction.
\newblock \emph{arXiv preprint arXiv:2010.06975}.

\bibitem[{Johnson et~al.(2016)Johnson, Pollard, Shen, Li-Wei, Feng, Ghassemi,
  Moody, Szolovits, Celi, and Mark}]{johnson2016mimic}
Alistair~EW Johnson, Tom~J Pollard, Lu~Shen, H~Lehman Li-Wei, Mengling Feng,
  Mohammad Ghassemi, Benjamin Moody, Peter Szolovits, Leo~Anthony Celi, and
  Roger~G Mark. 2016.
\newblock Mimic-iii, a freely accessible critical care database.
\newblock \emph{Scientific data}, 3(1):1--9.

\bibitem[{Lee et~al.(2020)Lee, Yoon, Kim, Kim, Kim, So, and
  Kang}]{lee2020biobert}
Jinhyuk Lee, Wonjin Yoon, Sungdong Kim, Donghyeon Kim, Sunkyu Kim, Chan~Ho So,
  and Jaewoo Kang. 2020.
\newblock Biobert: a pre-trained biomedical language representation model for
  biomedical text mining.
\newblock \emph{Bioinformatics}, 36(4):1234--1240.

\bibitem[{Li and Yu(2020)}]{li2020icd}
Fei Li and Hong Yu. 2020.
\newblock Icd coding from clinical text using multi-filter residual
  convolutional neural network.
\newblock In \emph{Proceedings of the AAAI Conference on Artificial
  Intelligence}, volume~34, pages 8180--8187.

\bibitem[{Lin et~al.(2017)Lin, Feng, Santos, Yu, Xiang, Zhou, and
  Bengio}]{lin2017structured}
Zhouhan Lin, Minwei Feng, Cicero Nogueira~dos Santos, Mo~Yu, Bing Xiang, Bowen
  Zhou, and Yoshua Bengio. 2017.
\newblock A structured self-attentive sentence embedding.
\newblock \emph{arXiv preprint arXiv:1703.03130}.

\bibitem[{May et~al.(2019)May, Wang, Bordia, Bowman, and
  Rudinger}]{may2019measuring}
Chandler May, Alex Wang, Shikha Bordia, Samuel~R Bowman, and Rachel Rudinger.
  2019.
\newblock On measuring social biases in sentence encoders.
\newblock \emph{arXiv preprint arXiv:1903.10561}.

\bibitem[{Mullenbach et~al.(2018)Mullenbach, Wiegreffe, Duke, Sun, and
  Eisenstein}]{mullenbach2018explainable}
James Mullenbach, Sarah Wiegreffe, Jon Duke, Jimeng Sun, and Jacob Eisenstein.
  2018.
\newblock Explainable prediction of medical codes from clinical text.
\newblock \emph{arXiv preprint arXiv:1802.05695}.

\bibitem[{Pascual et~al.(2021)Pascual, Luck, and
  Wattenhofer}]{pascual2021towards}
Damian Pascual, Sandro Luck, and Roger Wattenhofer. 2021.
\newblock Towards bert-based automatic icd coding: Limitations and
  opportunities.
\newblock \emph{arXiv preprint arXiv:2104.06709}.

\bibitem[{Reimers and Gurevych(2019)}]{reimers2019sentence}
Nils Reimers and Iryna Gurevych. 2019.
\newblock Sentence-bert: Sentence embeddings using siamese bert-networks.
\newblock In \emph{Proceedings of the 2019 Conference on Empirical Methods in
  Natural Language Processing and the 9th International Joint Conference on
  Natural Language Processing (EMNLP-IJCNLP)}, pages 3973--3983.

\bibitem[{Vaswani et~al.(2017)Vaswani, Shazeer, Parmar, Uszkoreit, Jones,
  Gomez, Kaiser, and Polosukhin}]{vaswani2017attention}
Ashish Vaswani, Noam Shazeer, Niki Parmar, Jakob Uszkoreit, Llion Jones,
  Aidan~N Gomez, Lukasz Kaiser, and Illia Polosukhin. 2017.
\newblock Attention is all you need.
\newblock In \emph{NIPS}.

\bibitem[{Vu et~al.(2020)Vu, Nguyen, and Nguyen}]{vu2020label}
Thanh Vu, Dat~Quoc Nguyen, and Anthony Nguyen. 2020.
\newblock A label attention model for icd coding from clinical text.
\newblock \emph{arXiv preprint arXiv:2007.06351}.

\bibitem[{Wang et~al.(2018)Wang, Li, Wang, Zhang, Shen, Zhang, Henao, and
  Carin}]{wang2018joint}
Guoyin Wang, Chunyuan Li, Wenlin Wang, Yizhe Zhang, Dinghan Shen, Xinyuan
  Zhang, Ricardo Henao, and Lawrence Carin. 2018.
\newblock Joint embedding of words and labels for text classification.
\newblock \emph{arXiv preprint arXiv:1805.04174}.

\bibitem[{Xie et~al.(2019)Xie, Xiong, Yu, and Zhu}]{xie2019ehr}
Xiancheng Xie, Yun Xiong, Philip~S Yu, and Yangyong Zhu. 2019.
\newblock Ehr coding with multi-scale feature attention and structured
  knowledge graph propagation.
\newblock In \emph{Proceedings of the 28th ACM International Conference on
  Information and Knowledge Management}, pages 649--658.

\bibitem[{Zhang et~al.(2019{\natexlab{a}})Zhang, Kishore, Wu, Weinberger, and
  Artzi}]{zhang2019bertscore}
Tianyi Zhang, Varsha Kishore, Felix Wu, Kilian~Q Weinberger, and Yoav Artzi.
  2019{\natexlab{a}}.
\newblock Bertscore: Evaluating text generation with bert.
\newblock \emph{arXiv preprint arXiv:1904.09675}.

\bibitem[{Zhang et~al.(2019{\natexlab{b}})Zhang, Wei, and
  Zhou}]{zhang2019hibert}
Xingxing Zhang, Furu Wei, and Ming Zhou. 2019{\natexlab{b}}.
\newblock Hibert: Document level pre-training of hierarchical bidirectional
  transformers for document summarization.
\newblock In \emph{Proceedings of the 57th Annual Meeting of the Association
  for Computational Linguistics}, pages 5059--5069.

\bibitem[{Zhang et~al.(2019{\natexlab{c}})Zhang, Chen, Yang, Lin, and
  Lu}]{zhang2019biowordvec}
Yijia Zhang, Qingyu Chen, Zhihao Yang, Hongfei Lin, and Zhiyong Lu.
  2019{\natexlab{c}}.
\newblock Biowordvec, improving biomedical word embeddings with subword
  information and mesh.
\newblock \emph{Scientific data}, 6(1):1--9.

\end{thebibliography}
\bibliographystyle{acl_natbib}

\appendix

\newpage
\section{Appendix}
\subsection{Document Segmentation}
Document segmentation is the very first step in the pre-processing pipeline. We leverage the pretrained spacy model "en\_core\_sci\_lg"\footnote{Available at \href{https://allenai.github.io/scispacy/}{https://allenai.github.io/scispacy/.} } that is trained specifically using biomedical text. To further remove the confusion brought by the bullet-like text such as "1. xx", "2. xx", we use regex to replace the "." with "," for better segmentation. Even by adding these rules, the segmentation is still making many mistakes caused by other text format issues such as sub-section titles and ambiguous line breaks. Exclusively enumerating fine-grained regex rules should help, more efforts should be invested in this direction.

\subsection{Model Averaging}
As indicated by Tab. \ref{tab_mimic_iii}, by simply averaging the predictions from different models, the performance boost at a large margin. We give more details of how the 3 models perform individually on the full MIMIC-III dataset. All these models are trained with different $w_l$ in Eq. \ref{eq_loss} and training strategies. 

\begin{table}[h]
  \centering
  \resizebox{\columnwidth}{!}{%
    \begin{tabular}{lc|cc|cc|cc}
    \hline
    \multirow{2}{2em}{Training\\Stage} &\multirow{2}{2em}{$w_l$ \\ Pair}  & \multicolumn{2}{c}{AUC-ROC} & \multicolumn{2}{c}{F1} & \multicolumn{2}{c}{P@K} \\ 
    & &  Macro & Micro & Macro & Micro & P@8 & P@15 \\ 
    single-stage & \{3, 10\}& 92.4	&98.8	&10.3	&55.6	&72.6	&57.6 \\ 
    single-stage & \{3, 5\}  & 91.9	&98.8	&9.9	&56.1	&73.2	&58.2 \\ 
    two-stage & \{3, 10\} & 93.2	&98.9	&10.1	&54.8	&72.1	&57.3  \\ \hline
    score-avg & -  & 92.5 &	98.9	&10.1	&55.5	&72.7	&57.7  \\ \hline
    model-avg & - & \textbf{94.2} & \textbf{99.2}   &	\textbf{10.4}	&	\textbf{57.6}	& \textbf{75.0} & \textbf{59.6}  \\ \hline
    \end{tabular}
    }
    \caption{We pick three models trained slightly differently. "single-stage" means we train the model without frozen the token transformer. The "score-avg" and "model-avg" are the ones reported in Tab. \ref{tab_mimic_iii}}.
  \label{tab_model_average}
\end{table}

When we dive deep into the performance, we found that these three models would perform well in different classes, this leads us to the model averaging technique. We strongly conjecture that this observation is caused by the relatively small dataset size at the document level. We only have fewer than 50k sequence of sentences to train the sentence transformer, making the sentence transformer trapped in bad local optima. 

\begin{table*}[t]
  \centering
  \resizebox{2\columnwidth}{!}{%
    \begin{tabular}{p{.2\linewidth}p{.8\linewidth}}
    \hline
    Sample 1 &  \\ \hline
    Title & Phase IV, single-center, randomized and open-label study to evaluate the impact of early treatment with a combination of an inhaled corticosteroid and a long-acting beta-2 adrenergic agonist (budesonide / formoterol) versus standard of care on the evolution of the disease in vulnerable patients with COVID-19: INHALAVID Trial \\ \hline
    Objective & To evaluate the impact of early treatment with a budesonide / formoterol versus standard of care on the evolution of the disease in vulnerable patients with COVID-19.\\ \hline
    Patient Population &  Vulnerable patients with COVID-19.\\ \hline
    Inclusion Criteria & 1. Patients capable of understanding the terms of the trial and agreeing to participate on it. 
2. Patients who sign the informed consent or oral consent with subsequent written confirmation. 
3. Patients > or = 65 years of age or > or = 50 years with a documented diagnosis in the medical history of at least one of the following comorbidities: arterial hypertension (diagnosed in the medical history and under treatment with antihypertensive drugs, cardiovascular disease, obesity (BMI > 30), diabetes, COPD, or active cancer. 
4. Positive antigen test for SARS-CoV-2. 
5. Less than 10 days from the appearance of the first symptoms at the time of randomization. \\ \hline \hline
    
    Sample 2 & \\ \hline
    Title & Fecal Microbiota Transplantation in the treatment of amyotrophic lateral sclerosis \\ \hline
    Objective & To establish the method of fecal microbiota transplantation for amyotrophic lateral sclerosis (ALS), to verify its efficacy and safety, and to study its potential mechanism, in order to provide new evidence and direction for the treatment and pathogenesis of ALS.\\ \hline
    Patient Population & Patients with Amyotrophic lateral sclerosis.\\ \hline
    Inclusion Criteria & (1)Patients were diagnosed as clinically definite ALS, clinically probable ALS, or clinically probable and laboratory supported ALS according to El Escorial criteria from the World Federation of Neurology in 1998. Patients who can't be defined as specific type or have family history were excluded. (2)According to the ALS or HSS, the disease was mild or moderate state; (3)Age ranges from 18 to 65; (4)Can cooperate to complete the inspection; (5)Agree to participate in this study and sign informed consent;\\ \hline
    
    
    \end{tabular}
    }
    \caption{Two TrailTrove Samples}.
  \label{tab_traitrove_samples}
\end{table*}

\subsection{Weight $w_l$ for Positive Samples}
We found the model training is sensitive to the $w_l$ in Eq. \ref{eq_weighted_CE}. In general, with a larger  $w_l$, the AUC-ROC scores (both Macro and Micro), as well as the F1-Macro score will increase (Tab. \ref{tab_model_average}), while the left scores go down. Our primary attempts show that we can easily boost the AUC-ROC and F1-Macro to SOTA at cost of the other three scores. As the community seems to weigh more on the F1-micro score \cite{vu2020label, mullenbach2018explainable}, we, therefore, did not explore the strategy of giving larger $w_l$ to infrequent classes. However, which score should be more important is discussable as in certain cases infrequent classes may need special attention. Nevertheless, $w_l$ provides an easy and explainable way to shift the focus among different classes.

\subsection{TF-IDF baseline}
We adopt the BM25 algorithm to calculate the similarity score of each query and document (SNOMED indication) pair $(q, i)$. We briefly describe the full pipeline of the algorithm in this section.

Let $M=3955$ be the total number of SNOMED indications in the search base. Let $W$ be the dictionary constructed from these indications. Let $q$ be a raw disease name (query) with length $N$ and $q_n \in W,\ n=1,\ldots,N$, be an $n$-th token in the query.  Let $\mathcal{I} = \{1,2,\ldots,M\}$, be the set of indices of all the indications in the search base. 

For each indication $i$ and each token $q_n$, we have the Frequency function $F: \mathcal{I} \times W \rightarrow \mathbb{N}$ defined as:

\begin{equation}
	F(i, q_n) = \frac{N(i, q_n)\cdot (1+k)} {N(i, q_n) + k \cdot \big( 1-b + b \cdot H(i)\big)},
\end{equation}
\begin{equation}
H(i) = \frac{N(i)}{\frac{1}{M} \sum_{i=1}^M N(i)},
\end{equation}
where $N(i, q_n)$ denote a number of times term $q_n$ appeared in indication $i$ and $N(i)$ is the total number of tokens that indication $i$ contains. We use $k=1.2$, $b=0.75$ in our paper.

Inverse Frequency $IF: W \rightarrow \mathbb{R}$ is defined as: 
\begin{equation}
	IF(q_n) = \log\bigg[1 + \frac{M - M(q_n) + 0.5}{M(q_n) + 0.5} \bigg],
\end{equation}
where $M(q_n)$ denotes the number of indications that contain the token $q_n$. Finally, the similarity between query $q$ and $i$-th indication is defined as
\begin{equation}
	s(q,i) = \sum_{n=1}^N \big[ F(i, q_n) \cdot IF(q_n) \big].
\end{equation}
The top1 hit picks the indication $i^{*}$ that has the maximum similarity score:
\begin{equation}
	i^{*} = \argmax_{i \in \mathcal{I}} s(q,i).
\end{equation}

\subsection{A Sample of TrailTrove}
We present two typical examples in this section (Tab. \ref{tab_traitrove_samples}) to show how TrialTrove data appear. Note that we apply no text segmentation is applied to the trial title field. Text from different fields is joined by the token "[SEP]" before it is fed to the model.

\end{document}